\documentclass[10pt,twocolumn,letterpaper]{article}

\usepackage{cvpr}              
\definecolor{cvprblue}{rgb}{0.21,0.49,0.74}
\usepackage[pagebackref,breaklinks,colorlinks,allcolors=cvprblue]{hyperref}

\usepackage{algorithm}
\usepackage{listings}
\usepackage{xcolor}

\lstdefinestyle{python}{
  language=Python,
  basicstyle=\ttfamily\small,
  keywordstyle=\color{blue},
  stringstyle=\color{red!70!black},
  commentstyle=\color{gray},
  showstringspaces=false,  
  showspaces=false,        
  keepspaces=true,         
  frame=tb,
  breaklines=true,
  tabsize=2,
}

\usepackage{amsmath}
\usepackage{amsfonts}
\usepackage{bm}

\def\paperID{*****} 
\def\confName{CVPR}
\def\confYear{2026}

\title{E-MMDiT: Revisiting Multimodal Diffusion Transformer Design for Fast Image Synthesis under Limited Resources}

\author{
Tong Shen \quad
Jingai Yu \quad
Dong Zhou \quad
Dong Li\quad
Emad Barsoum \\[0.3em]
Advanced Micro Devices, Inc.\\
{\tt\small \{T.Shen, Jingai.Yu, Dong.Zhou, d.li, Emad.Barsoum\}@amd.com}
}

\begin{document}
\maketitle

\renewcommand{\arraystretch}{1.2}

\begin{abstract}

Diffusion models have shown strong capabilities in generating high-quality images from text prompts. However, these models often require large-scale training data and significant computational resources to train, or suffer from heavy structure with high latency. To this end, we propose Efficient Multimodal Diffusion Transformer (E-MMDiT), an efficient and lightweight multimodal diffusion model with only 304M parameters for fast image synthesis requiring low training resources. 
We provide an easily reproducible baseline with competitive results.
Our model for 512px generation, trained with only 25M public data in 1.5 days on a single node of 8 AMD MI300X GPUs, achieves 0.66 on GenEval and easily reaches to 0.72 with some post-training techniques such as GRPO.
Our design philosophy centers on \textbf{token reduction} as the computational cost scales significantly with the token count. We adopt a highly compressive visual tokenizer to produce a more compact representation and propose a novel multi-path compression module for further compression of tokens. To enhance our design, we introduce \textbf{Position Reinforcement}, which strengthens positional information to maintain spatial coherence, and \textbf{Alternating Subregion Attention (ASA)}, which performs attention within subregions to further reduce computational cost. In addition, we propose \textbf{AdaLN-affine}, an efficient lightweight module for computing modulation parameters in transformer blocks. 
Our code is available at \url{https://github.com/AMD-AGI/Nitro-E} and we hope E-MMDiT serves as a strong and practical baseline for future research and contributes to democratization of generative AI models.
\end{abstract}

\begin{figure}[t]
\centering
\includegraphics[width=1\columnwidth]{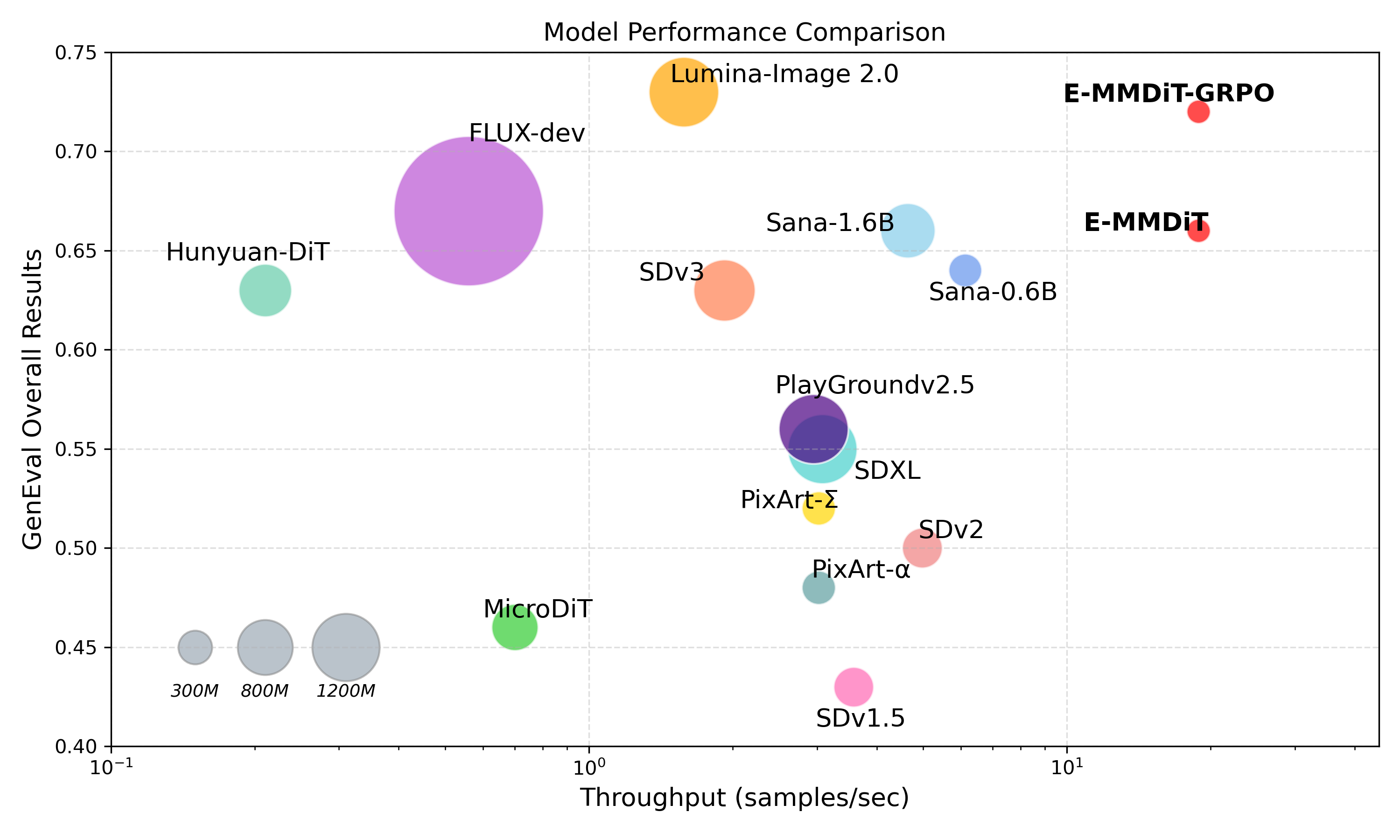} 
\caption{Comparison with other models on GenEval and throughput. 
Throughput is measured by generating 512px images using a batch size of 32 and 20 steps on an AMD MI300X GPU.
Despite having only 304M parameters, our model achieves competitive GenEval performance and a clear advantage in throughput.}
\label{fig_comparison}
\end{figure}

\section{Introduction}

\begin{figure*}[t]
  \centering
  \begin{subfigure}[b]{0.198\textwidth}
    \includegraphics[width=\linewidth]{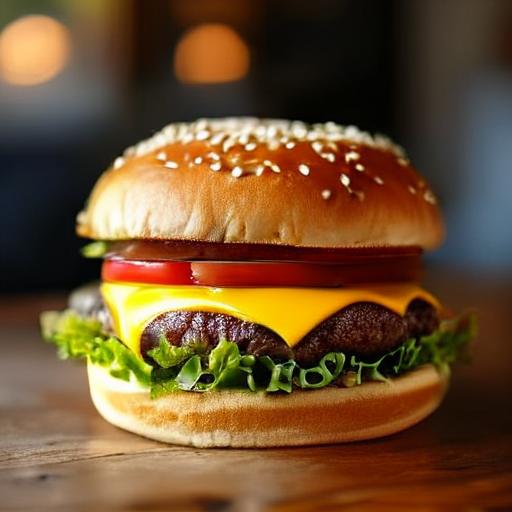}
  \end{subfigure}
  \begin{subfigure}[b]{0.198\textwidth}
    \includegraphics[width=\linewidth]{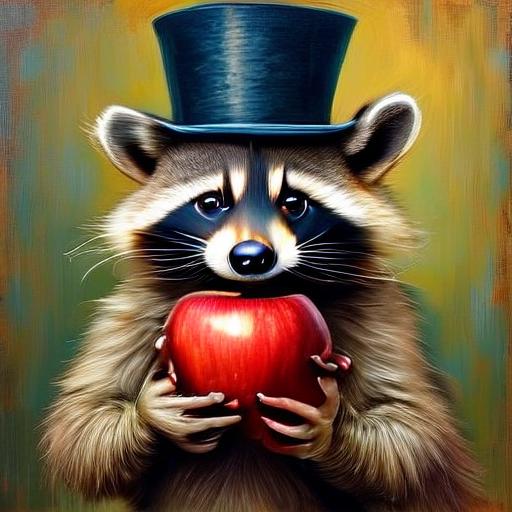}
  \end{subfigure}
  \begin{subfigure}[b]{0.198\textwidth}
    \includegraphics[width=\linewidth]{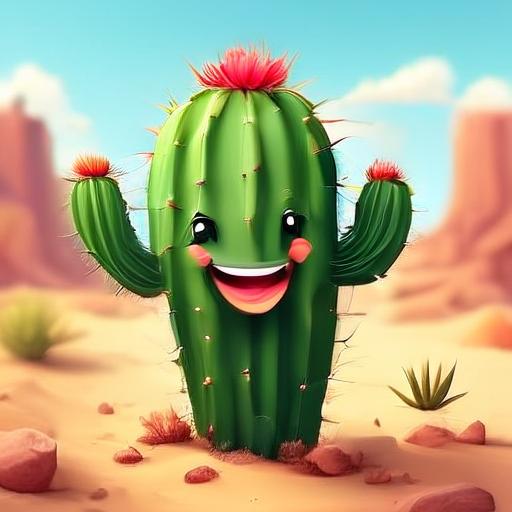}
  \end{subfigure}
  \begin{subfigure}[b]{0.198\textwidth}
    \includegraphics[width=\linewidth]{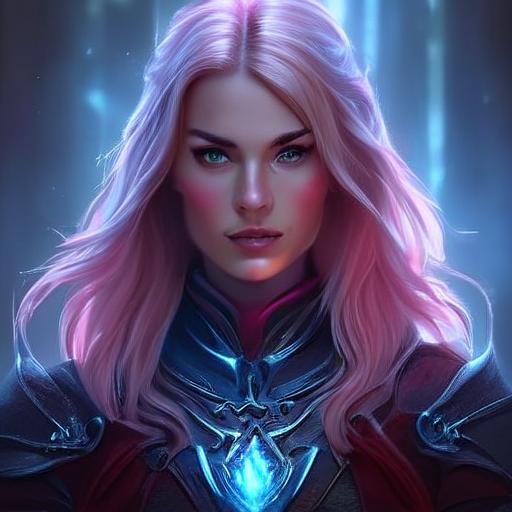}
  \end{subfigure}

  \vspace{0.2em}

  \begin{subfigure}[b]{0.4\textwidth}
    \includegraphics[width=\linewidth]{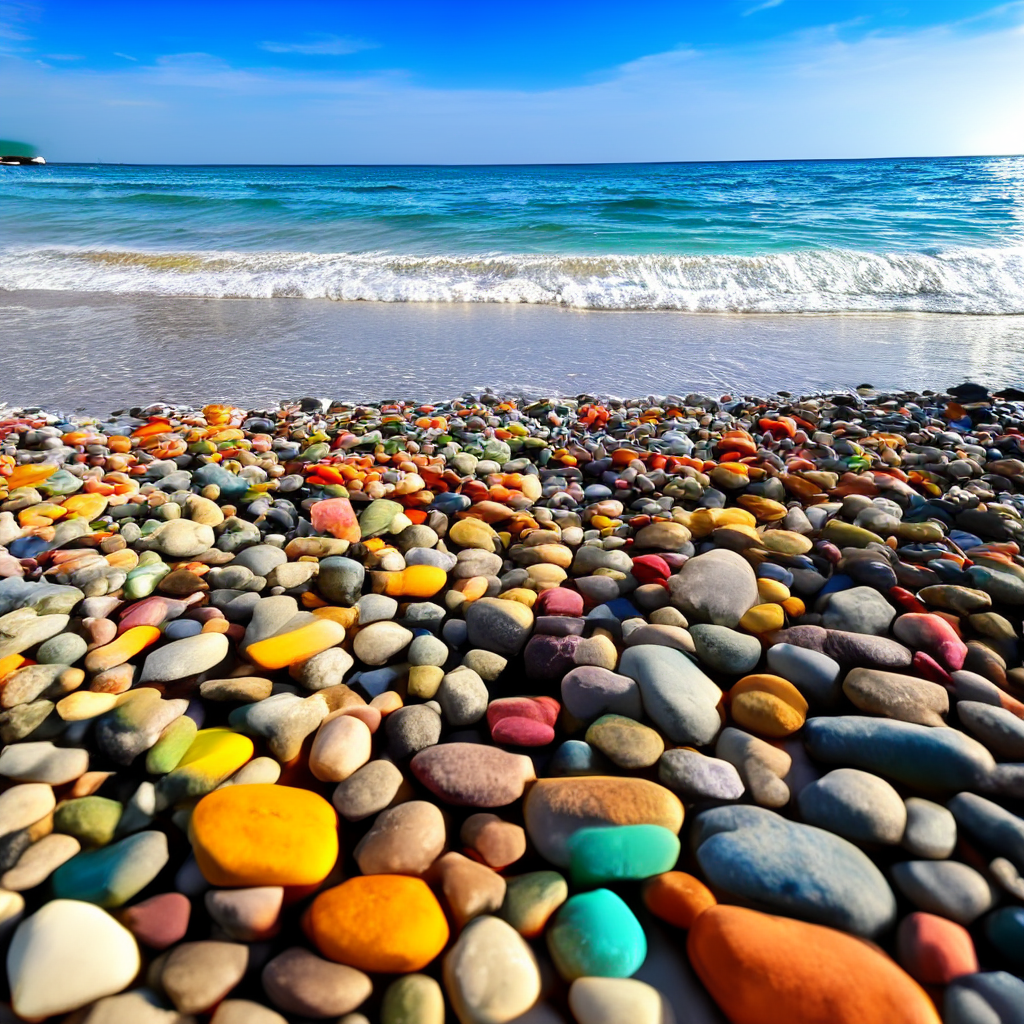}
  \end{subfigure}
  \begin{subfigure}[b]{0.4\textwidth}
    \includegraphics[width=\linewidth]{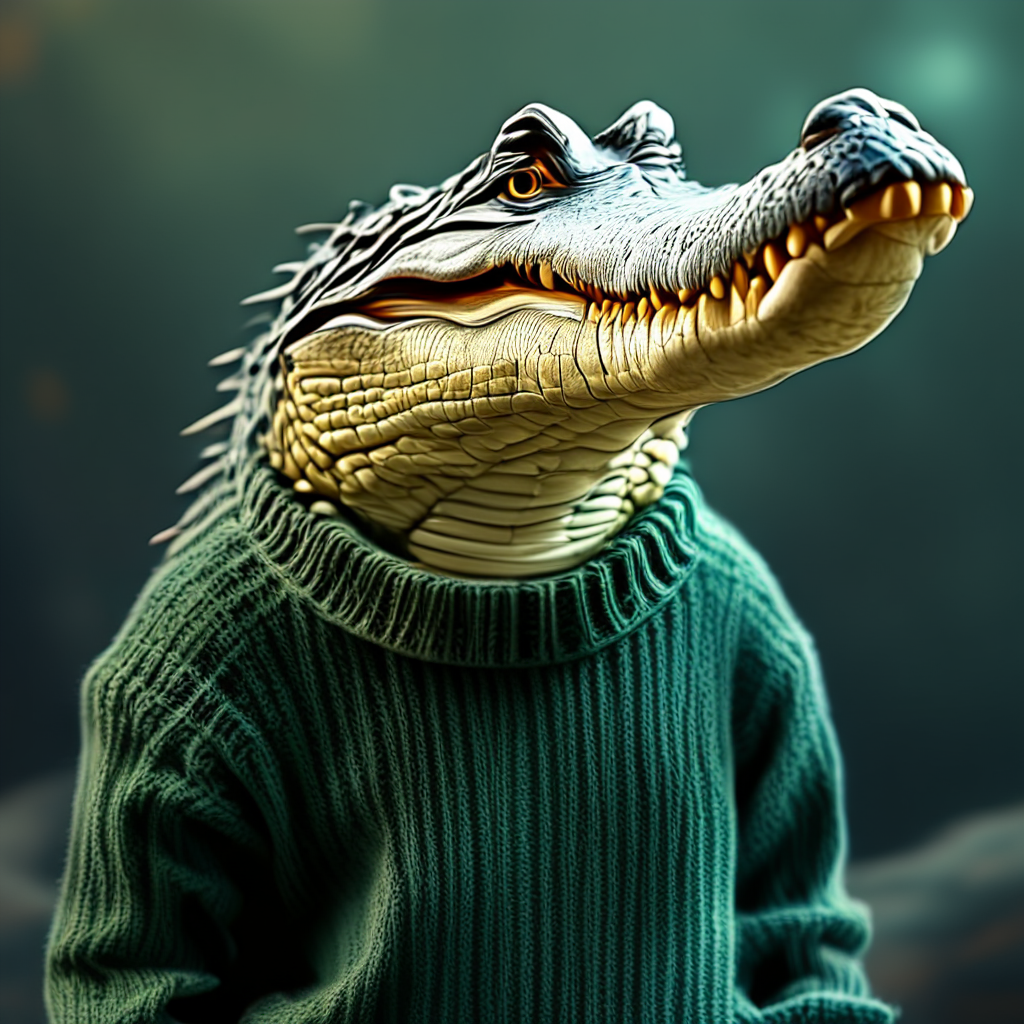}
  \end{subfigure}
  \caption{Images generated by our 304M E-MMDiT model at 512px (top) and 1024px (bottom). }
  \label{fig:grid}
  
\end{figure*}

In recent few years, large-scale text-to-image diffusion models \cite{pixart, sd, sd3, sdxl} have achieved great success, also enabling a wide range of applications, such as controllable generation \cite{controlnet}, personalization \cite{textual_inv}, and video generation \cite{imagenvideo}. 
However, these models suffer from slow inference due to their iterative generation and often contain a large number of parameters, which becomes a barrier for deployment.

To improve efficiency and deployment-friendliness of these models, various compression techniques are commonly adopted, including pruning \cite{structural_pruning} to reduce the number of parameters and quantization \cite{qdiffusion} to accelerate computations by simplifying operators. Another direction involves distillation methods \cite{dmd, ladd} that aim to significantly reduce the number of inference steps and thus lower cost.

While these compression techniques are applicable to the diffusion models, they work as an add-on solution rather than addressing the core model design. Therefore, searching for efficient and deployment-friendly base models remains crucial. Given the limited availability of such models and high training cost of diffusion models, we believe it is still worthy to explore design of light-weight models especially those with low-cost training. 

In this paper, we discuss this problem by proposing E-MMDiT, an efficient diffusion model with only 304M parameters. Our model shows competitive generation abilities with high throughput (Refer to Figure \ref{fig_comparison}).
Our E-MMDiT builds upon MMDiT \cite{sd3}, a transformer-based architecture with separate weights for different modalities. Compared with vanilla Diffusion Transformer (DiT) \cite{dit}, MMDiT offers a more unified framework for handling diverse input types, making it a promising framework to build upon.

We begin our design by focusing on \textbf{token reduction}, which is a key aspect addressed by existing works \cite{sana, microdiff}. For example, \cite{sana} argues that the visual tokenizer should take full responsibility for compression and adopts a highly compressive tokenizer to significantly reduce tokens involved. 
In another work, \cite{microdiff} targets low-budget training by introducing a ``deferred masking'' strategy during training, which drops a large proportion of tokens after being encoded by a ``patch mixer'', demonstrating the redundancy of tokens. This concept is further supported by \cite{tokenshuffle}, which discusses token compression techniques. Building upon these insights, we combine the merits of both approaches by adopting a highly compressive visual tokenizer and proposing a novel multi-path compression module to further reduce tokens during the forward pass. 

Building on our novel multi-path token compression module, we propose several additional components to enhance efficiency and effectiveness of E-MMDiT.
\textbf{Position Reinforcement} addresses the weakening of spatial cues caused by token compression and restores spatial coherence by reattaching positional information to the reconstructed tokens.
\textbf{Alternating Subregion Attention (ASA)} offers a computationally efficient alternative to full self-attention by dividing tokens into subregions and performing attention independently in a parallel manner. Unlike prior work UDiT \cite{udit} that suffers from limited inter-group communication and requires extra spatial depthwise convolutional layers, ASA dynamically alternates the grouping strategy, enabling effective inter-region interactions without extra components. 
\textbf{AdaLN-affine} computes modulation parameters for transformer blocks by producing affine transformations of a global vector, avoiding requirement of block-specific MLPs, thus significantly reducing parameters and overhead.

Our contributions are summarized as:
\begin{itemize}
    \item We present an efficient diffusion model E-MMDiT with only 304M parameters for fast image synthesis under limited training and inference resources.
    
    \item We propose a collection of novel designs and conduct comprehensive experiments to validate our designs.
    
    \item We showcase how a light-weight diffusion model can be trained from scratch in 1.5 days on a single node of 8 AMD MI300X GPUs with only 25M public data. Our model, which achieves competitive performance on four widely used metrics, is easily reproducible and thus can server as a strong baseline for future research of the field.

\end{itemize}

\section{Related Work}

Image generation is a fundamental and widely studied task in the field of Generative AI. In the early stage, Adversarial Generative Networks (GANs) \cite{gan} played an important role in GAN-based image generation methods \cite{stylegan, biggan, stylegant}. Another family of methods \cite{llamagen, var, mar}, built upon auto-regression, have also achieved remarkable progress.

Diffusion models, another line of research, have emerged as a dominant paradigm in the field.
Denoising Diffusion Probabilistic Models (DDPMs) \cite{ddpm} and Denoising Diffusion Implicit Models (DDIMs) \cite{ddim} are early fundamental works of diffusion models that provide theoretical formulation of diffusion models. Later on, diffusion models are applied to large-scale text-to-image generation tasks with various model sizes \cite{pixart, sd, sdxl, flux, sana}. The architecture has also shifted from U-Net \cite{unet}, to DiT \cite{dit} and MMDiT \cite{sd3}. 

To accelerate diffusion models, different approaches have been explored, such as quantization \cite{qdiffusion, accptq}, caching techniques \cite{cachemeifyoucan, learningtocache}, pruning \cite{structural_pruning, ldpruner}, and step distillation \cite{dmd, ladd, ufogen}. 
In this paper, we focus on the model architecture itself by proposing several novel designs and training our model from scratch with much lower training cost. 
\section{Design of E-MMDiT}

\begin{figure}[t]
\centering
\includegraphics[width=0.9\columnwidth]{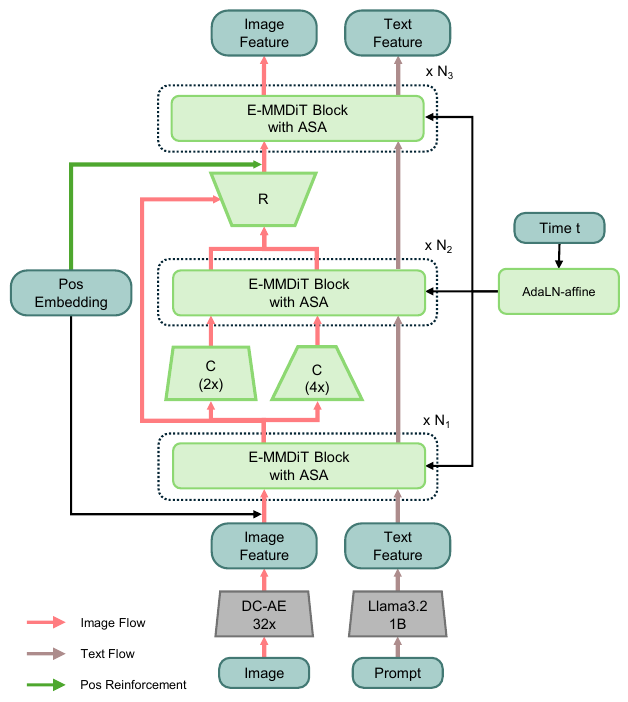} 
\caption{Illustration of E-MMDiT. Image is encoded by a highly compressive tokenizer DC-AE (ratio of 32$\times$), and prompt is encoded by a light-weight LLM, Llama3.2-1B. Our E-MMDiT blocks are incorporated with ASA for faster token interaction.
After first $N_1$ blocks, the tokens are further condensed by our multi-path compression module with ratio of 2$\times$ and 4$\times$ for the following $N_2$ blocks. The tokens are finally recovered by the token reconstructor and processed by the final $N_3$ blocks. Positional Embedding is additionally added to the reconstructed tokens for position reinforcement. AdaLN-affine encodes timestep and provides modulation parameters for each block through an affine transformation of the global vector. }
\label{fig_emmdit}
\end{figure}


\subsection{Diffusion Architectures}

U-Net \cite{unet} was widely adopted as the base architecture for some early diffusion models \cite{ddim, sd}. U-Net is a convolution-based model with downsampling and upsampling blocks, as well as skip connection for feature merging. For text-to-image generation, text embeddings are integrated via cross-attention layers in each block. 

While convolution-based models are known to introduce spatial locality bias that is suitable for modeling images, Vision Transformers (ViTs) \cite{vit} has marked a paradigm shift in almost all vision tasks, including diffusion models, known as Diffusion Transformers (DiTs) \cite{dit}.
Studies show that the inductive bias of U-Net is not essential for image generation. DiTs have demonstrated promising performance and better scalability, making them a popular alternative \cite{sana, pixart}.

To further unify text and image features, a variant of DiT called MMDiT has been proposed \cite{sd3, flux}. 
MMDiT processes different modalities (e.g. text and image features) using separate sets of weights, and inter-modality interaction is achieved by a joint attention mechanism over concatenated features, replacing the typical combination of self-attention and cross-attention.
MMDiT offers a more unified framework for handling different modalities, making a promising base model to build on.

Figure \ref{fig_emmdit} is a simplified illustration of our proposed E-MMDiT model. For text features, we follow a recent trend of replacing the cumbersome text-encoder T5 \cite{t5} with some light-weight Large Language Model (LLM) \cite{sana}. Specifically, we choose Llama 3.2-1B \cite{llama3}, which is much lighter than T5 with 4.7B parameters. For image tokenizer, in line with our design principle of \textbf{Token Reduction}, we adopt a highly compressive tokenizer, DC-AE \cite{dcae}, for a more compact representation. 
In addition, we propose a novel multi-path compression module to further reduce the number of tokens involved in the diffusion process. 
Our module compresses tokens with two different ratios, 2$\times$ and 4$\times$, and processes both token sets jointly. 
There are three groups of E-MMDiT blocks, with block numbers $(N_1, N_2, N_3)$. The middle group operates on the compressed tokens while the other two groups process the tokens in the original resolution.
The compressed tokens are recovered by a token reconstructor and fed to the third block group for prediction.
To preserve spatial consistency, positional embeddings are injected into both the initial and reconstructed tokens, a design we refer to as \textbf{Position Reinforcement}. 
\textbf{AdaLN-affine} encodes global information, i.e. timestep, and provides modulation parameters for each block by an affine transformation of a global vector.
Each E-MMDiT block incorporates an \textbf{Alternating Subregion Attention (ASA)} module, which serves as a lightweight and effective alternative to full self-attention.

In the next sections, we discuss each design in detail.

\begin{algorithm}[t]
\caption{Python code for Subregion Division.}
\label{listing_regdivision}

\begin{lstlisting}[style=python]
from einops import rearrange
x = rearrange(x, 
    "b (l s n) c -> (b s) (l n) c", 
    n=chunk_size, s=region_num)
\end{lstlisting}

\end{algorithm}

\subsection{Token Reduction}

In transformers, the training cost heavily depends on the number of tokens, as self-attention has a quadratic complexity of $O(N^2)$ where $N$ is the number of tokens involved. Therefore, reducing tokens is a straightforward strategy to improve efficiency.
Tokens are compact representations defined in a latent space instead of raw pixels. For DiT models such as PixArt \cite{pixart} and SDv3 \cite{sd3}, images are first compressed by a factor of 8 and then patchified by a patch size of 2, resulting in an effective downsampling ratio of 16$\times$. Recent work \cite{sana} argues that the visual tokenizer should take full responsibility for compression and leave the transformer solely for denoising. Following this principle, they adopt a highly compressive visual tokenizer, DC-AE, which has an aggressive down-sampling factor of 32. Without further patchification, This results in 32$\times$ downsampling ratio and a 75\% reduction in token count. 
In E-MMDiT, we take advantage of DC-AE as well and at the same time, propose a novel multi-path compression module for further token condensing.

Visual information is highly redundant, despite latent encoding. 
MicroDiT \cite{microdiff} demonstrates this by randomly dropping a large proportion of tokens during training using a ``deferred masking'' strategy, while still successfully training the model. Instead of dropping tokens, our multi-path compression module compress tokens using two ratios, 2$\times$ and 4$\times$, producing two sets of tokens that are jointly processed by the following blocks. 
Compared with MicroDiT that applies dropping ratio from 50\% up to 75\%, 
our method achieves a comparable level of token reduction (68.5\%).
More importantly, unlike ``deferred masking'' that is used only during training, our token compression is effective during both training and inference. 

The compression and reconstruction modules are inspired by TokenShuffle \cite{tokenshuffle}. The compressor merges locally adjacent tokens along the channel dimension and processes them using a small Multi-Layer Perceptron (MLP). The reconstructor recovers the original token resolution by reversing the compression process, untangling the tokens through the channel dimension with an MLP. Additionally, a skip connection from early blocks is added to aid information recovery during reconstruction.

Our E-MMDiT blocks are divided into three groups indicated by $(N_1, N_2, N_3)$ in Figure \ref{fig_emmdit}. We need a few blocks working on the original resolution, $N_1$ and $N_3$, but most of the blocks $N_2$ are used to process the compressed tokens.

\subsection{Position Reinforcement}

Positional Embedding (PE) is important to transformers, as it informs each token of its spatial location in an image. In this work, we follow the setting in SDv3 \cite{sd3} and adopt absolute PE. Specifically PE is constructed using sine and cosine functions at different frequencies and injected to token embeddings at the input stage.

As we mentioned in the previous section, our approach involves token compression and recovery, which might weaken or distort the original positional information. 
To alleviate this issue, we propose to explicitly reinforce positional information on the reconstructed tokens. As depicted in Figure \ref{fig_emmdit}, PE is applied in both the input and reconstruction stage. We show in the experiments that this strategy helps maintain spatial coherence and improve performance.

\begin{figure}[t]
\centering
\includegraphics[width=0.9\columnwidth]{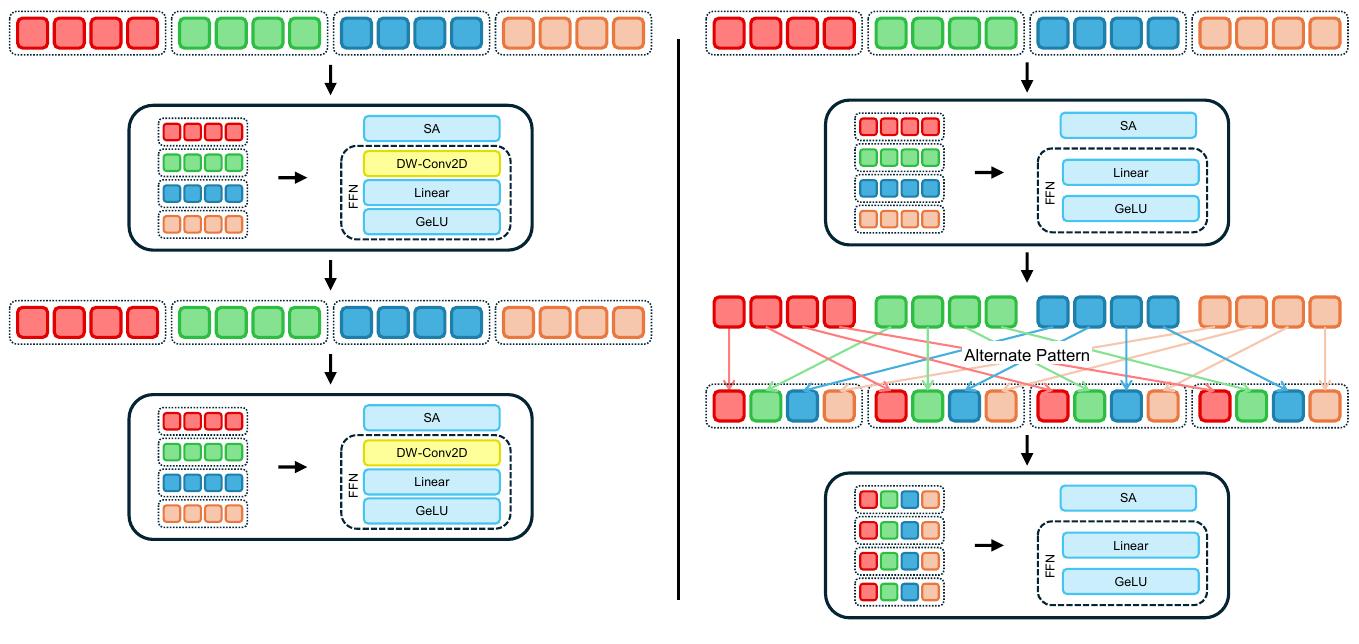} 
\caption{Illustration of ASA with two consecutive blocks. Tokens are represented as one dimensional sequences for simplicity. 
Left side depicts the downsampled attention in UDiT, where tokens are always divided by the same group pattern, lacking inter-group communication. Extra depthwise Convolutions are required in the FFN.
In contrast, our proposed ASA shown on the right simply alternates grouping patterns for the second block. 
It is easy to observe that tokens grouped by the same color in the first block are reorganized into groups containing tokens of different colors, thus enabling interaction across subregions.
}
\label{fig_subattn}
\end{figure}

\subsection{AdaLN-affine}
As analyzed in \cite{pixart}, the linear projections used in the Adaptive LayerNorm (AdaLN) layers to compute the modulation parameters account for a substantial proportion of the total parameters. These modulation parameters, denoted by $S^{(i)}$, are obtained by a block-specific MLP. To save computation and reduce parameters, they propose AdaLN-single, in which $\bar{S}$ is a global vector shared across all blocks. The block-specific parameters are then computed by $S^{(i)}=\hat{S}+\beta^{(i)}$, where $\beta^{(i)}$, being as a bias term, is a trainable vector maintained by each DiT block.

We further improve the flexibility by proposing AdaLN-affine, where a scale term $\gamma^{(i)}$ is learned as well for each block, modifying the formulation to $S^{(i)}=\hat{S}(1+\gamma^{(i)})+\beta^{(i)}$. This formulation applies both scale and bias to the global vector, making it an affine transformation. 


\begin{table*}[t]
\centering
\small
\resizebox{0.9\linewidth}{!}{
\begin{tabular}{cccccccccc}
\toprule
Methods                          & \begin{tabular}[c]{@{}c@{}}Throughput\\ (samples/s)\end{tabular} & \begin{tabular}[c]{@{}c@{}}Latency\\ (ms)\end{tabular} & \begin{tabular}[c]{@{}c@{}}Parameters\\ ~Main network\\ ~(M)\end{tabular} & \begin{tabular}[c]{@{}c@{}}TFLOPs\\ Main network~\end{tabular} & \begin{tabular}[c]{@{}c@{}}Dataset Size\\ (M)\end{tabular} & GenEval\(\scriptstyle\uparrow\) & IR\(\scriptstyle\uparrow\)   & HPS\(\scriptstyle\uparrow\)   & DPG\(\scriptstyle\uparrow\)    \\ 
\hline
\multicolumn{10}{c}{Large-scale Models (1024px)}                                                                                                                                                                                                                                                                                                                                                                  \\ 
\hline
Hunyuan-DiT \cite{hunyuandit}                     & 0.21                                                             & 5356                                                   & \textbf{1500}                                                                      & 14.37                                                          & -                                                          & 0.63    & 0.92 & 30.22 & 78.90  \\
FLUX-dev \cite{flux}                        & 0.56                                                             & 2943                                                   & 11901                                                                     & 21.50                                                          & -                                                          & \underline{0.67}    & 0.82 & \textbf{32.47} & 84.00  \\
Sana-1.6B \cite{sana}                       & 1.34                                                             & \textbf{705}                                                    & \underline{1604}                                                                      & 2.95                                                           & 50                                                         & 0.66    & \underline{0.99} & 27.76 & \underline{84.80}  \\
Lumina-Image 2.0 \cite{gao2024lumina-next}               & 1.58                                                             & 1243                                                   & 2610                                                                      & 4.99                                                           & 110                                                        & \textbf{0.73}    & 0.69 & 29.53 & \textbf{87.20}  \\
SDv3 \cite{sd3}                            & 1.92                                                             & \underline{819}                                                    & 2028                                                                      & 2.11                                                           & 1000                                                       & 0.63    & 0.87 & 31.53 & 84.10  \\

PlayGroundv2.5 \cite{playground}                  & \underline{2.95}                                                             & 1126                                                   & 2567                                                                      & \textbf{1.58}                                                           & -                                                          & 0.56    & \textbf{1.09} & \underline{32.38} & 75.50  \\ 
SDXL \cite{sdxl}                            & \textbf{3.08}                                                             & 1036                                                   & 2567                                                                      & \underline{1.59}                                                           & -                                                          & 0.55    & 0.69 & 30.64 & 74.70  \\
\hline
\multicolumn{10}{c}{Light-weight
  Models (512px)}                                                                                                                                                                                                                                                                                                                                                         \\ 
\hline
MicroDiT \cite{microdiff}                        & 0.70                                                             & 1849                                                   & 1160                                                                      & 1.13                                                           & 37                                                         & 0.46    & 0.81 & 27.23 & 72.90  \\
PixArt-$\Sigma$ \cite{pixartsigma} & 3.02                                                             & 625                                                    & 610                                                                       & 1.24                                                           & 33                                                         & 0.52    & \textbf{0.97} & \textbf{30.37} & 80.50  \\
PixArt-$\alpha$ \cite{pixart} & 3.02                                                             & 625                                                    & 610                                                                       & 1.24                                                           & 25                                                         & 0.48    & 0.92 & \underline{29.95} & 71.60  \\
SDv1.5 \cite{sd}                           & 3.58                                                             & 642                                                    & 860                                                                       & 0.80                                                           & 2000                                                       & 0.43    & 0.19 & 24.24 & 63.18  \\
SDv2 \cite{sd}                            & 4.98                                                             & 498                                                    & 866                                                                       & 0.80                                                           & 3900                                                       & 0.50    & 0.29 & 26.38 & 64.20  \\

Sana-0.6B \cite{sana}                       & \underline{6.13}                                                             & \underline{424}                                                    & \underline{592}                                                                       & \underline{0.32}                                                           & 50                                                         & 0.64    & \underline{0.93} & 27.22 & \textbf{84.30}  \\
E-MMDiT-512                      & \textbf{18.83}                                                            & \textbf{398}                                                    & \textbf{304}                                                                       & \textbf{0.08}                                                           & 25                                                         & \underline{0.66}    & \textbf{0.97} & 29.82 & 81.60  \\
E-MMDiT-512-GRPO                 & \textbf{18.83}                                                            & \textbf{398}                                                    & \textbf{304}                                                                       & \textbf{0.08}                                                           & 25                                                         & \textbf{0.72}    & \textbf{0.97} & 29.82 & \underline{82.04}  \\ 
\hline
\multicolumn{10}{c}{Light-weight
  Models (1024px)}                                                                                                                                                                                                                                                                                                                                                        \\ 
\hline
PixArt-$\Sigma$ \cite{pixartsigma} & 0.52                                                             & 2363                                                    & 610                                                                       & 6.50                                                           & 33                                                         & 0.54    & 0.87 & 30.05 & 80.50  \\
PixArt-$\alpha$ \cite{pixart} & 0.54                                                             & 2184                                                   & 610                                                                       & 6.50                                                           & 25                                                         & 0.47    & 0.94 & \textbf{30.68} & 71.69  \\
Sana-0.6B \cite{sana}                       & \underline{1.88}                                                             & \underline{707}                                                    & \underline{592}                                                                       & \underline{1.12}                                                           & 50                                                         & 0.64    & 0.97 & 27.71 & \textbf{83.60}  \\
E-MMDiT-1024                     & \textbf{5.54}                                                             & \textbf{432}                                                    & \textbf{304}                                                                       & \textbf{0.25}                                                           & 14                                                         & \underline{0.66}    & \underline{0.98} & 30.16 & 82.35  \\
E-MMDiT-1024-GRPO                & \textbf{5.54}                                                             & \textbf{432}                                                    & \textbf{304}                                                                       & \textbf{0.25}                                                           & 14                                                         & \textbf{0.71}    & \textbf{1.00} & \underline{30.23} & \underline{82.39}   \\
\bottomrule
\end{tabular}
}

\caption{Comparison of our E-MMDIT with other SOTA models. Throughput and Latency are tested on a AMD MI300X GPU with FP16 precision. Latency: End-to-end cost measured with batch=1 and sampling step=20 for generating an image. Throughput: Measured with batch=32 and sampling step=20.
TFLOPs: Calculated for one forward pass of the diffusion model. 
Throughput and Latency are averaged results over multiple runs.
Models are grouped based on their model size and latency to ensure fair comparison on similar levels. We highlight the \textbf{best}, \underline{second best} for each group. All evaluations use official implementations without any additional optimizations.
}
\label{table_modelcomparison}
\end{table*}

\subsection{Alternating Subregion Attention (ASA)}

The attention mechanism, as the fundamental operation in transformers, plays a core role in enabling interactions between token pairs. However, due to its quadratic complexity $O(N^2)$ with respect to the token count $N$, its computational cost grows quickly as the token increases.
Approaches have been proposed to reduce the cost of attention. For instance, Sana \cite{sana} replaces self-attention with ReLU-based linear attention, reducing the complexity to $O(N)$. While this significantly lowers computation, the absence of non-linear similarity function may lead to sub-optimal performance, as stated in the paper. 
To this end, they introduce depthwise convolutions and Gated Linear Units in the Feedforward Network (FFN) to capture more information. 

UDiT \cite{udit} provides another perspective to optimize attention. They divide the tokens into four 2$\times$ downsampled groups, and apply self-attention to the groups in parallel, reducing the computation to $1/4$ of full attention. However, this design restricts interactions within individual groups, limiting inter-group interactions. They mitigate this by incorporating 2D and depthwise convolution in their FFN.

Our model introduces Alternating Subregion Attention (ASA), a simple yet effective design.
Following UDiT's idea of parallel token groups, we use a flexible grouping strategy with different patterns per block, avoiding extra layers and preventing attention from being restricted to fixed regions, which proves effective in practice.


To better understand how ASA works, we show a simplified toy example with two consecutive transformer blocks in Figure \ref{fig_subattn}. For simplicity, tokens are represented as one dimensional sequences. 
The left side depicts the downsampled attention used in UDiT, where tokens are always divided by the same group pattern.
To enable interactions between groups, their design incorporates a modified FFN that includes multiple depthwise convolutions with various kernel sizes.
In contrast, our proposed ASA shown on the right has different group division strategies across blocks.
We observe that in the first block, the groups are formed by tokens in the same color, while in the second block, the grouping strategy is alternated and the groups contain tokens of different colors. 
This alternating grouping strategy achieves an effective ``receptive field'' equivalent to full attention over time, without requiring additional components.

Given a token sequence \textbf{x} with shape $(B, L, C)$, representing batch number, sequence length, and channel width respectively. Our region division is formally defined by two parameters, represented by a tuple (\texttt{region\_num}, \texttt{chunk\_size}), used to specify the number of subregions and the size of token chunk formed by consecutive tokens. The region division is implemented as in Algorithm \ref{listing_regdivision}.

\subsection{Training Objectives}
\textbf{Rectified Flow} Our model is trained with Rectified Flows \cite{flowmatching} that defines the forward process as a straight path between data distribution and noise, as in $\textbf{x}_t=(1-\sigma_t)\textbf{x}_0+\sigma_t\bm{\epsilon}$, where $t$ represents timestep and $\sigma_t$ is a timestep-dependent variable controlled by a scheduler; $\textbf{x}_0$ is the image and $\bm{\epsilon}$ denotes noise from standard Gaussian Distribution $\mathcal{N}(0, I)$. The diffusion objective is defined as predicting velocity formed by image and noise:

\begin{equation}
    \mathcal{L}_\text{RF}(\bm{\theta}) := \mathbb{E}_{\bm{\epsilon}\sim\mathcal{N}(0, I), t}\| (\bm{\epsilon}-\textbf{x}_0)-v_{\bm{\theta}}(\textbf{x}_t,t) \|_2^2,
\end{equation}

where $v_{\bm{\theta}}(\cdot)$ represents our diffusion model parameterized by $\bm{\theta}$ that predicts velocity of the noised input $\textbf{x}_t$.

\textbf{Representation Alignment Loss} It has been explored that aligning features of the diffusion model with a pre-trained visual encoder helps accelerate convergence \cite{repa}. We use REPresentation Alignment (REPA) loss as a regularizer. The loss is defined as:

\begin{equation}
    \mathcal{L}_\text{REPA}(\bm{\theta}, \bm{\phi}) := -\mathbb{E}_{\textbf{x}_0,t,\bm{\epsilon}}\text{sim}(g(\textbf{x}_0),h_{\bm{\phi}}(f_{\bm{\theta}}(\textbf{x}_t))),
\end{equation}

where $\text{sim}(\cdot,\cdot)$ is a similarity function; $g(\cdot)$ is the visual encoder, e.g. DINO v2 \cite{dinov2}; $h_{\bm{\phi}}(\cdot)$ is a projection head that maps features of the diffusion model $f_{\bm{\theta}}(\textbf{x}_t)$.

Final objective is defined as:
\begin{equation}
    \mathcal{L} := \mathcal{L}_\text{RF} + \lambda\mathcal{L}_\text{REPA}, 
\end{equation}

where $\lambda$ is the weighting parameter.
\section{Experiments}

\begin{table}[t]
\centering
\small
\resizebox{0.9\linewidth}{!}{
\begin{tabular}{ccccccc}
\toprule
Model             & Tput  & GenEval\(\scriptstyle\uparrow\) & IR\(\scriptstyle\uparrow\)   & HPS\(\scriptstyle\uparrow\)   & DPG\(\scriptstyle\uparrow\)    \\ 
\hline
512px       & 18.83                                                                                                                & 0.66    & 0.97 & 29.82 & \textbf{81.60}  \\
512px-dist  & \textbf{39.36}                                                                                                                 & \textbf{0.67}    & \textbf{0.99} & \textbf{30.18} & 78.77  \\
\hline
1024px      & 5.54                                                                                                                 & \textbf{0.66}    & 0.98 & 30.16 & \textbf{82.35}  \\
1024px-dist & \textbf{11.7}                                                                                                                 & 0.65    & \textbf{1.00} & \textbf{31.18} & 79.04   \\
\bottomrule
\end{tabular}
}
\caption{Performance of the distilled models. 
Compared with the original full-step models, the distilled models double the throughput and maintain similar performance across all metrics.  }
\label{table:dist}
\end{table}

\begin{figure}[t]
  \centering
  \setlength{\tabcolsep}{2pt} 
  \renewcommand{\arraystretch}{0.9} 
  \begin{tabular}{cc}
    \textbf{20-step} & \textbf{4-step} \\[0.3em]
    \includegraphics[width=0.22\textwidth]{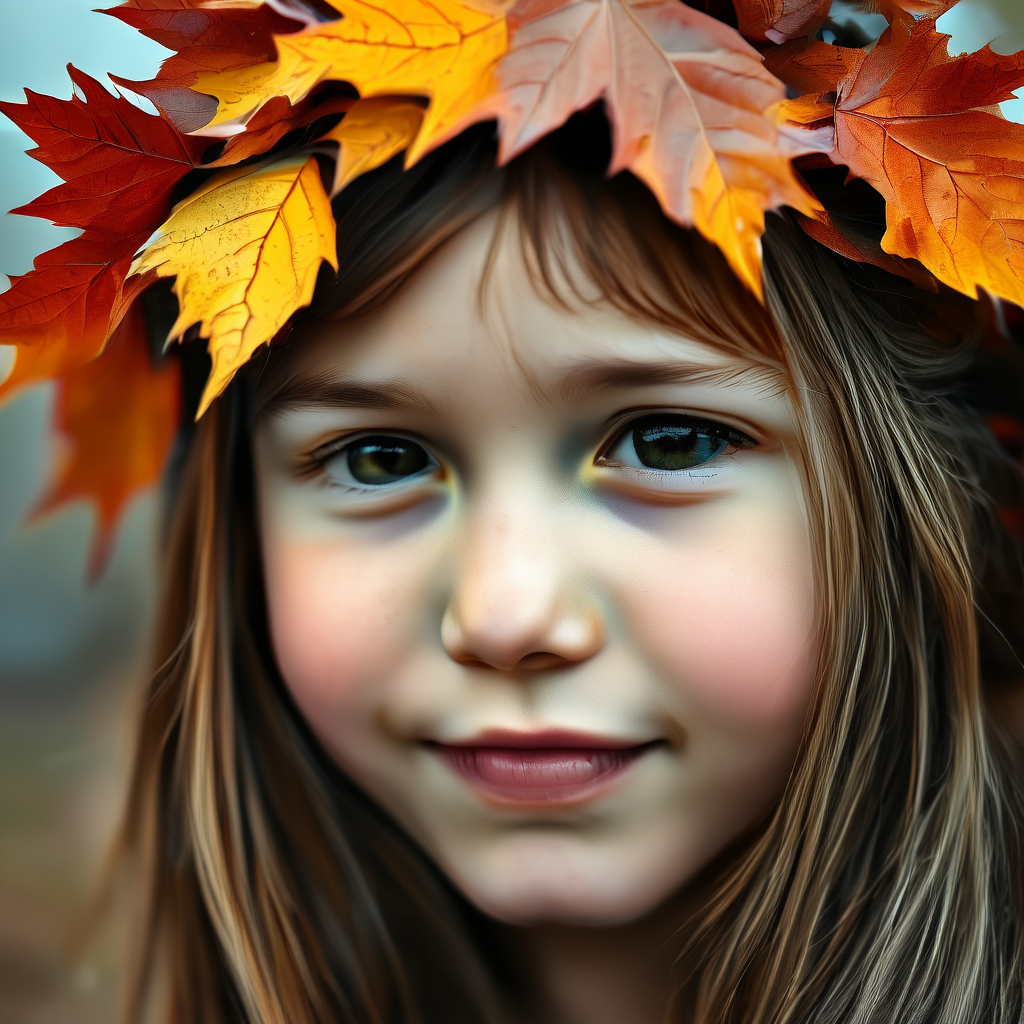} &
    \includegraphics[width=0.22\textwidth]{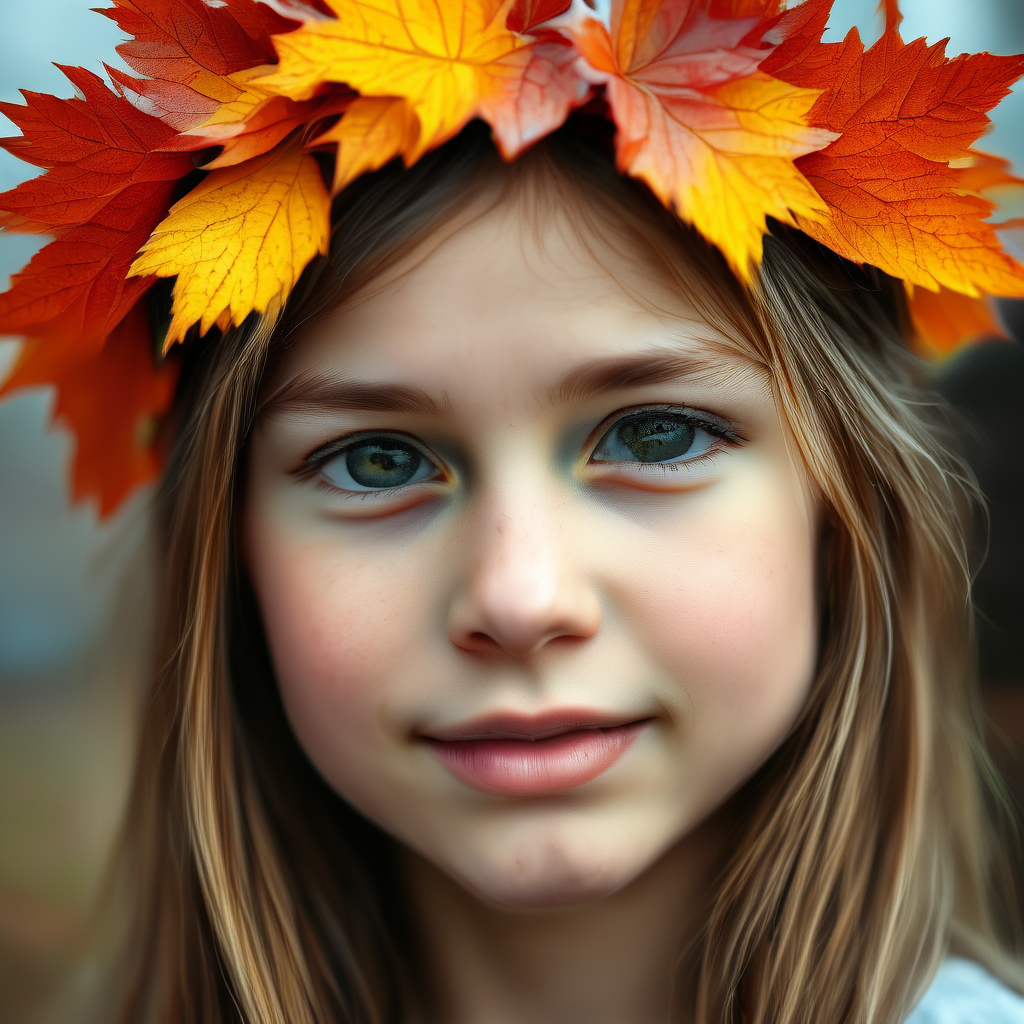} \\[0.2em]
    \includegraphics[width=0.22\textwidth]{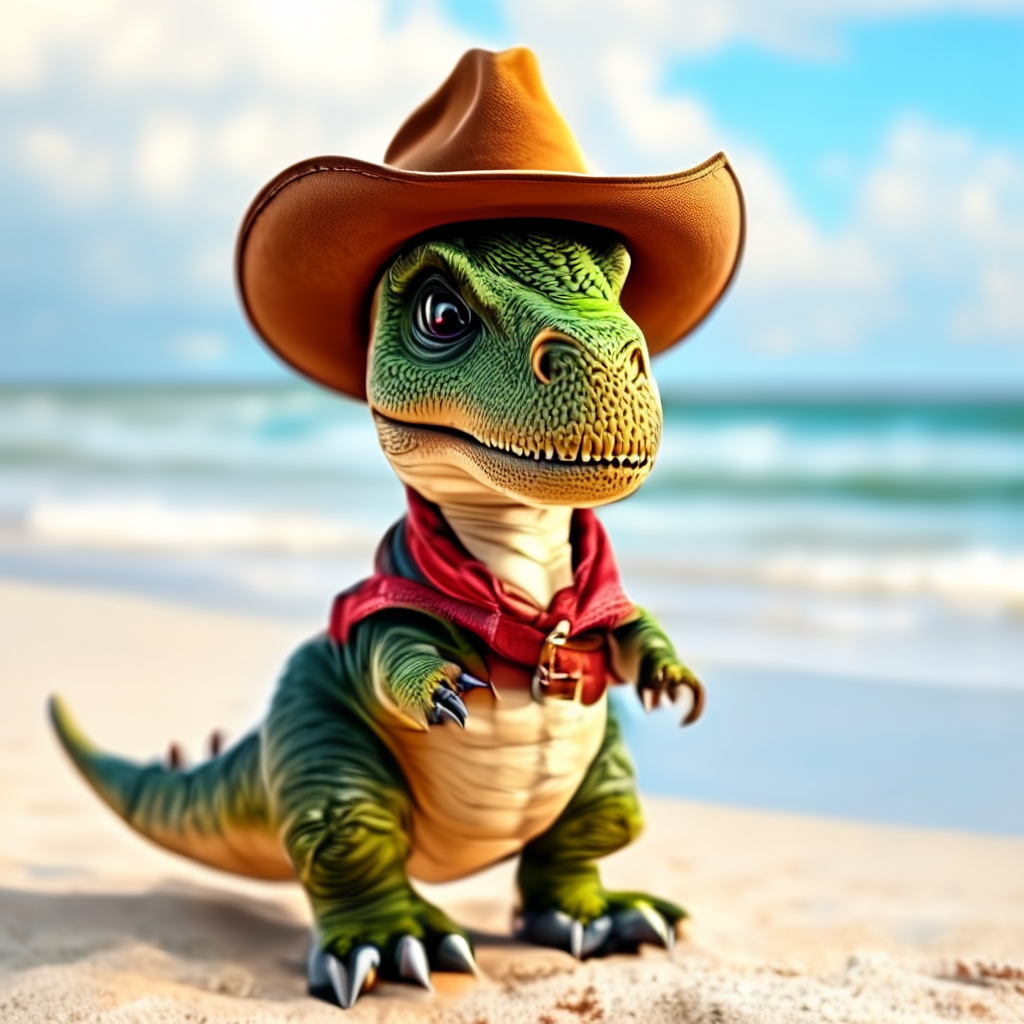} &
    \includegraphics[width=0.22\textwidth]{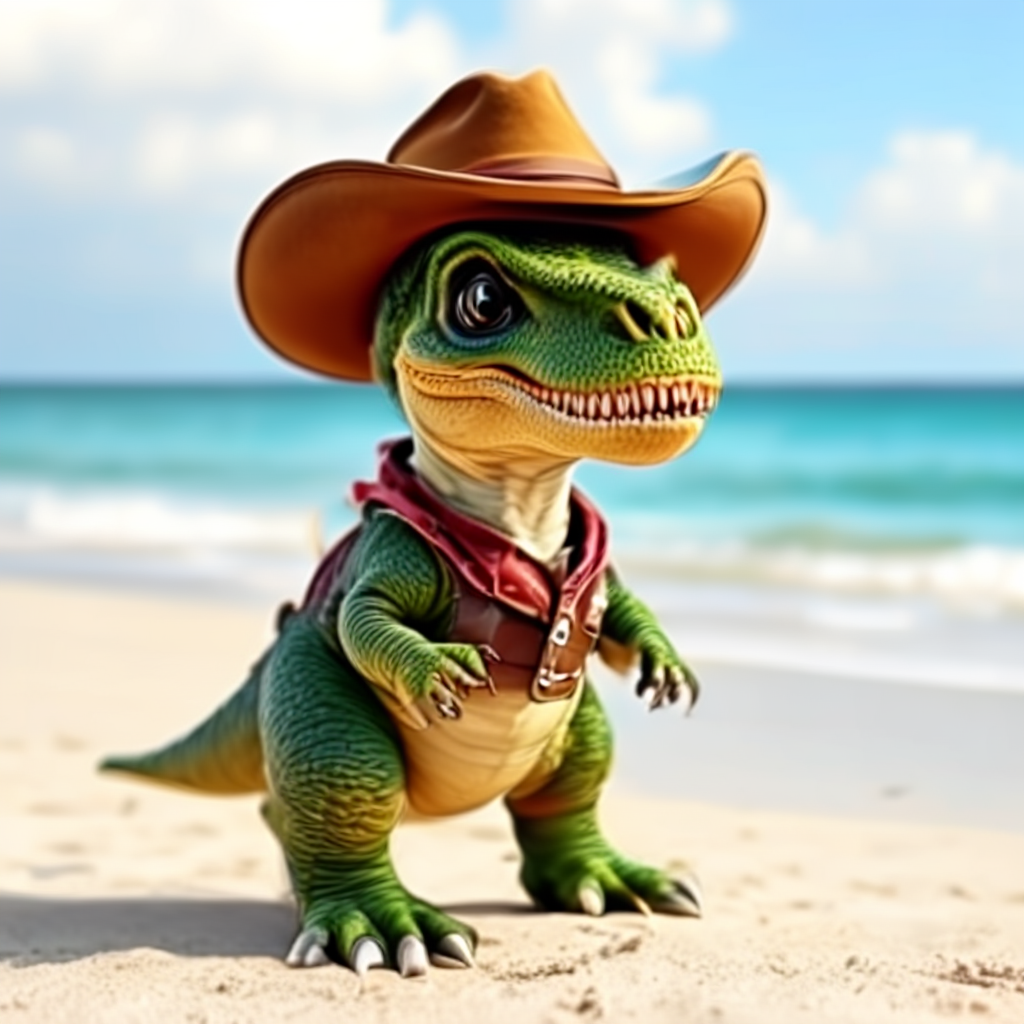} \\
  \end{tabular}
  \caption{Visual comparison between the distilled and the full-step models. The 4-step results maintain the same visual quality as the original 20-step results.}
  \label{fig:dist}
\end{figure}

\subsection{Model Details}
Our E-MMDiT consist of 24 Transformer blocks, each with 24 attention heads and 32 channels per head. The group numbers $[N_1, N_2, N_3]$ are set to $[4, 16, 4]$. The FFN has a multiplier of 3 instead of 4 for reduced parameters. 
For ASA, we set every three blocks as a group with parameters $[(1, 1), (4, 1), (4, 4)]$, where it is a full-attention blocks followed with two subregion attention blocks.
The token compressors are small MLP with two Linear layers and a GELU layer. The token reconstructor has three similar MLPs, two for upsampling and one for fusing tokens.

\subsection{Training Details}

\subsubsection{Dataset}

To make results easily reproducible, all our experiments are conducted on public data without any internal data. For text-to-image generation, we adopt a combination of real and synthetic data, resulting in totally 25M text-image pairs:

\begin{itemize}
    \item \textbf{SA1B} \cite{sam} comprises 11.1M high quality real-world images, originally used for segmentation tasks. We use the generated captions from \cite{pixart}.
    \item \textbf{JourneyDB} \cite{journeydb} contains 4.4M synthetic text-image pairs collected from Midjourney. 
    \item \textbf{FLUXDB} is another synthetic dataset we constructed using FLUX.1 model. The dataset contains 9.5M generated images whose prompts are collected from DataComp-1B \cite{datacomp1b} and DiffusionDB \cite{diffuiondb}.
\end{itemize}

For ablation studies, we conduct relatively small-scale experiments on ImageNet \cite{imagenet}.

\subsubsection{Optimization parameters}

Our text-to-image model is trained at resolution of both 512px and 1024px. We apply a two-stage training strategy using AdamW optimizer \cite{adamw} with batch size of 2048 on 8 AMD Instinct MI300X GPUs. Image and text features are pre-computed to accelerating training. An additional post-training stage with Group Relative Policy Optimization (GRPO) is optionally applied. The model can be further distilled into a faster model supporting few-step generation using adversarial distillation technology \cite{ladd}.

\begin{itemize}
    \item \textbf{Stage1}. We train our model for 100k iterations on full data with a learning rate of 3e-4. The REPA loss is applied during this stage to accelerate convergence, where features from DINO-v2 are used as the alignment target. This stage only applies to 512px resolution.
    \item \textbf{Stage2}. The SA1B dataset, despite its high quality, still contains intentionally obscured regions with blur or mosaic for privacy purposes. So in this stage, we finetune our model solely on the synthetic data for 50k iterations with 512px or 1024px resolutions. We enable Exponential Moving Average (EMA) for more stable convergence and omit the REPA loss.
    \item \textbf{Post-training} is an optional stage where we enhance our model using GRPO for 2k iterations with a combination reward of GenEval and HPSv2.1.
    \item  \textbf{Step Distillation} We distill our models following open-source project ``Nitro-1" \cite{nitro1}, where we generate 1M synthetic data using the teacher model and distill the full-step model into a few-step version supporting 1-4 steps.

\end{itemize}

For ablation experiments, we train a class-to-image model on ImageNet at a resolution of 256px. We use DiT-L/2 as our base model incorporating all of our proposed designs. Following the original setup in \cite{dit}, we train the model for 400k iterations without applying extra regularization (e.g. REPA) for fair comparison.

\begin{table}[t]
\centering
\resizebox{0.75\linewidth}{!}{
\begin{tabular}{lllll}
\toprule

Model            & \begin{tabular}[c]{@{}c@{}}FLOPs\\(G)\end{tabular} & \begin{tabular}[c]{@{}c@{}}Params\\(M)\end{tabular} & FID   & IS     \\ 
\hline
DiT L/2          & 161.42                                                & 458                                                 & 23.33 & 58.18  \\
Two-branch & 89.77                                                 & 343                                               & \textbf{22.42} & \textbf{58.65}  \\
w/o skip & 89.23                                                 & 342                                               & 28.75 & 48.16  \\
2$\times$ only & 81.58                                                  & 323                                               & 23.78 & 56.03  \\
4$\times$ only & 61.93                                                 & 336                                               & 33.52 & 41.43  \\
Stacked 2$\times$ & 73.56                                                 & 333                                               & 24.22 & 54.99  \\
\bottomrule

\end{tabular}
}
\caption{Ablation on different compression strategies. We compare different settings, two-branch with and without skip connection, single branch with only 2$\times$ or 4$\times$, or a stacked 2$\times$ design similar to UNet.}
\label{tab:ds}
\end{table}

\subsubsection{Metrics}

For text-to-image generation, we evaluate our model on four widely used metrics, GenEval \cite{geneval}, HPSv2.1 \cite{hpsv2}, DPG-Bench \cite{dpg} and ImageReward (IR) \cite{imagereward}. GenEval and DPG-Bench measure text-image alignment, while HPSv2.1 and ImageReward assess human preferences.
For class-to-image experiments on ImageNet, we use two common metrics, Fréchet Inception Distance (FID) and Inception Score (IS), to evaluate generation quality and diversity.

\subsubsection{Results}
We compare our models with various open-sourced models, including 
SDXL \cite{sdxl}, SDv3 \cite{sd3}, Sana-1.6B, Sana-0.6B \cite{sana}, MicroDiT \cite{microdiff}, FLUX-dev \cite{flux}, Lumina-Image 2.0 \cite{gao2024lumina-next}
, HunyuanDiT \cite{hunyuandit}, PlayGroundv2.5 \cite{playground}, SDv1.5, SDv2 \cite{sd}, PixArt-$\Sigma$ \cite{pixartsigma} and PixArt-$\alpha$ \cite{pixart}, shown in Table \ref{table_modelcomparison}.
The models are grouped based on their model size and FLOPs for fair comparison on similar computational levels.
Our model achieves competitive scores in the metrics, ranking highest on GenEval and ImageReward and delivering comparable results on HPS and DPG. More importantly, our model exhibits a clear advantage in terms of inference cost. Our model, with only 304M parameters, achieves the lowest latency among all candidates. Moreover, with larger batch size, it demonstrates a strong throughput advantage, outperforming other models by a wide margin, thanks to the extremely low FLOPs of the main network. 

To further speed up inference, our distilled models achieve twice the throughput while maintaining comparable metric scores as shown in Table \ref{table:dist}. Visual results illustrated in Figure \ref{fig:dist} further confirm this, demonstrating an effective solution for edge deployment.

\begin{table}[t]
\centering
\resizebox{0.75\linewidth}{!}{
\begin{tabular}{lllll}
\toprule

Model            & \begin{tabular}[c]{@{}c@{}}FLOPs\\(G)\end{tabular} & \begin{tabular}[c]{@{}c@{}}Params\\(M)\end{tabular} & FID   & IS     \\
\hline
DiT L/2          & 161.42                                                & 458                                                 & 23.33 & 58.18  \\
   (4, 16, 4)   & 89.77                                                 & 343                                               & \textbf{22.42} & \textbf{58.65}  \\
      (2, 20, 2) & 71.45          & 343           & 29.34                     & 45.85  \\
      (0, 24, 0) & 53.31          & 343           & 44.99                     & 30.18  \\
      (8, 8, 8) & 126.05          & 343           & 23.47                     & 55.40  \\
\bottomrule

\end{tabular}
}
\caption{Ablation on block configurations. The tuple indicates block number for each group, as in $(N_1, N_2, N_3)$.}
\label{tab_blks}
\end{table}

\begin{table}[t]
\centering
\resizebox{0.7\linewidth}{!}{
\begin{tabular}{lllll}
\toprule

Model            & \begin{tabular}[c]{@{}c@{}}FLOPs\\(G)\end{tabular} & \begin{tabular}[c]{@{}c@{}}Params\\(M)\end{tabular} & FID   & IS     \\
\hline
   PR\_R & 89.77                                                 & 343                                               & \textbf{22.42} & \textbf{58.65}  \\
      w/o PR & 89.77          & 343           & 24.78                     & 53.85  \\
      PR\_C & 89.77          & 343           & 26.56                     & 51.23  \\
      PR\_CR & 89.77          & 343           & 23.92                     & 54,94  \\
\bottomrule

\end{tabular}
}
\caption{Ablation on Position Reinforcement (``PR''). The suffix ``C'' and ``R'' represent Compressed and Reconstructed tokens respectively, indicating if we apply PR to them.}

\label{tab_pos}
\end{table}

\begin{table}[t]
\centering
\resizebox{0.65\linewidth}{!}{
\begin{tabular}{llll}
\toprule

Setting            & \begin{tabular}[c]{@{}c@{}}FLOPs\\(G)\end{tabular} & FID   & IS     \\
\hline
    w/o ASA & 12.9                    &  23.33 &  58.18 \\
   (1:1, 4:1, 4:4) & 6.4                                                                                               & \textbf{23.50} & \textbf{59.40}  \\
      (4:1, 4:4, 1:1) & 6.4                    & 24.55                     & 57.88  \\
      (4:1, 4:4) & 3.2                 & 26.54                     & 55.16  \\
      (4:1, 1:1, 4:4) & 6.4                     & 24.69                     & 57.17  \\
\bottomrule

\end{tabular}
}
\caption{Ablation on ASA.}
\label{tab_asa}
\end{table}

\subsection{Ablation Study}
We choose to use a more standard benchmark for ablation studies to further validate our designs, which is ImageNet $256 \times 256$ generation. We apply all of our designs to the model DiT-L/2 and evaluate the effectiveness.

\subsubsection{Downsampling Strategy}
We compare different strategies for token compression in Table \ref{tab:ds}. Our model with the novel two-branch compression module clearly outperforms other configurations in FID and IS, such as 2$\times$ only and 4$\times$ only or Stacked 2$\times$ similar to a UNet structure. 
Compared with original DiT L/2, our design has much 25\% less parameters and 44\% less FLOPs. In addition, the setting without skip connection obtains much worse scores, showing the importance of low-level features for reconstructing tokens.

\subsubsection{Block Configuration}
We conduct experiments with different block settings for $[N_1, N_2, N_3]$, shown in Table \ref{tab_blks}. We observe that when more blocks are assigned to process compressed tokens, although the computational cost is consistently reduced, it does not always lead to better performance. (0, 24, 0) is the extreme case that works similarly as a patchifier, which has the worst performance. Our final design (4, 16, 4) strikes a good balance between quality and efficiency.

\subsubsection{Position Reinforcement}
We also explore effectiveness of Position Reinforcement in Table \ref{tab_pos}. The suffix ``C'' and ``R'' represent compressed and reconstructed tokens respectively. It is interesting to observe that Position Reinforcement works better when only being applied to the reconstructed tokens. Reinforcing the compressed tokens even brings negative effect. 
\subsubsection{ASA Module}
We explore effectiveness of ASA module by experimenting various configurations. To further highlight the reduction in attention-related overhead, we isolate the FLOPs associated with the attention shown in Table \ref{tab_asa}. 
Each tuple indicates an ASA grouping. When set to $1:1$, ASA reduces to a standard full attention. It is evident that ASA significantly reduces FLOPs. Only adopting subregion attention, $(4:1, 4:4)$, saves the most computation, but at the cost of decreased quality. Our proposed design, one full attention block followed by two subregion attention blocks, cuts FLOPs by half while achieving slightly better results.
It is also worth noting that even with the same computational cost, the order of these blocks matters, demonstrated by the other two configurations.

\subsubsection{AdaLN-Affine}
We also study effectiveness of AdaLN-Affine in Table \ref{tab_adaln}. Both AdaLN-Single and AdaLN-Affine help reduce parameters and FLOPs of the original DiT baseline. 
Compared with AdaLN-Single, AdaLN-Affine improves both FID and IS with negligible overhead, which is not even reflected.

\begin{table}[t]
\centering
\resizebox{0.8\linewidth}{!}{
\begin{tabular}{lllll}
\toprule

Model            & \begin{tabular}[c]{@{}c@{}}FLOPs\\(G)\end{tabular} & \begin{tabular}[c]{@{}c@{}}Params\\(M)\end{tabular} & FID   & IS     \\
\hline
    DiT L/2 & 161.42                                                & 458                                                 & 23.33 & 58.18  \\
   AdaLN-Single & 89.77          & 343           & 22.94 & 56.60   \\
      AdaLN-Affine & 89.77                                                 & 343                                               & \textbf{22.42} & \textbf{58.65}  \\
\bottomrule

\end{tabular}
}
\caption{Ablation on AdaLN-Affine.}
\label{tab_adaln}
\end{table}
\section{Conclusion}

In this paper, we explore the design of efficient diffusion models with low computational cost in both training and inference. We introduce E-MMDiT, a lightweight MMDiT-based transformer with only 304M parameters.
Our core design philosophy emphasizes token reduction: we leverage a highly compressive visual tokenizer and propose a novel multi-path token compression module. To further improve performance, we incorporate three enhancements: Position Reinforcement, Alternating Subregion Attention (ASA), and AdaLN-Affine.
E-MMDiT achieves competitive scores on four widely used benchmarks, while exhibiting a strong advantage in throughput, outperforming other models by a large margin.
We have released our code with all details, hoping this serves as a strong baseline and encourages future work in efficient visual generation.

{
    \small
    \bibliographystyle{ieeenat_fullname}
    \bibliography{main}
}


\end{document}